\documentclass[lettersize,journal]{IEEEtran}
\usepackage{amsmath,amsfonts}
\usepackage{algorithmic}
\usepackage{algorithm}
\usepackage{array}
\usepackage[caption=false,font=normalsize,labelfont=sf,textfont=sf]{subfig}
\usepackage{textcomp}
\usepackage{stfloats}
\usepackage{url}
\usepackage{booktabs}
\usepackage{verbatim}
\usepackage{graphicx}
\usepackage{xcolor}

\usepackage[colorlinks=true, linkcolor=blue, citecolor=blue, filecolor=blue, urlcolor=blue]{hyperref}  

\usepackage{cleveref}
\usepackage{cite}
\usepackage{multirow}
\usepackage{tabularx}

\crefname{figure}{Fig.}{Figs.}  

\begin{document}

\title{AFANet: Adaptive Frequency-Aware Network for Weakly-Supervised Few-Shot Semantic Segmentation}

\author{Jiaqi Ma,
        Guo-Sen Xie,
        Fang Zhao,
        Zechao Li,~\IEEEmembership{Senior Member,~IEEE}
\thanks{Manuscript received 3 June 2024; revised 6 September 2024; accepted 18 October
2024.(\textit{Corresponding authors: Guo-Sen Xie and Zechao Li.})} 
\thanks{Jiaqi Ma, Guo-Sen Xie, and Zechao Li are with the School of Computer Science and Engineering, Nanjing University of Science and Technology, Nanjing 210094, China. E-mail: machiachi@163.com, gsxiehm@gmail.com, zechao.li@njust.edu.cn.}
\thanks{Fang Zhao is with the School of Intelligence Science and Technology, Nanjing University, Suzhou 215163, China. E-mail: fzhao@nju.edu.cn.}}

\markboth{IEEE TRANSACTIONS ON MULTIMEDIA}%
{Shell \MakeLowercase{\textit{et al.}}: A Sample Article Using IEEEtran.cls for IEEE Journals}


\maketitle
\begin{abstract}
Few-shot learning aims to recognize novel concepts by leveraging prior knowledge learned from a few samples. However, for visually intensive tasks such as few-shot semantic segmentation, pixel-level annotations are time-consuming and costly. Therefore, in this paper, we utilize the more challenging image-level annotations and propose an adaptive frequency-aware network (AFANet) for weakly-supervised few-shot semantic segmentation (WFSS). Specifically, we first propose a cross-granularity frequency-aware module (CFM) that decouples RGB images into high-frequency and low-frequency distributions and further optimizes semantic structural information by realigning them. Unlike most existing WFSS methods using the textual information from the multi-modal language-vision model, e.g., CLIP, in an offline learning manner, we further propose a CLIP-guided spatial-adapter module (CSM), which performs spatial domain adaptive transformation on textual information through online learning, thus providing enriched cross-modal semantic information for CFM. Extensive experiments on the Pascal-5\textsuperscript{i} and COCO-20\textsuperscript{i} datasets demonstrate that AFANet has achieved state-of-the-art performance. The code is available at https://github.com/jarch-ma/AFANet.
\end{abstract}


\section{Introduction}
\IEEEPARstart{O}{ver} the past decade, deep learning methods have achieved remarkable performance in image segmentation tasks \cite{citation28},\cite{citation29},\cite{citation30},\cite{citation31}. However, most of these methods require a large amount of data for training, and the process of collecting data and manual annotation is time-consuming and costly. Previous statistics have shown that for a single 1280×720 pixel image, the time required for annotation is approximately 1.5 hours \cite{ref1}. The cost of annotating massive datasets has become increasingly difficult to bear, particularly in cases where domain expertise is required, highlighting the issue of data scarcity. To address these challenges, weakly-supervised semantic segmentation (WSSS) and few-shot semantic segmentation (FSS) have been proposed and play a significant role in various fields such as medical image analysis \cite{ref2}, autonomous driving \cite{ref3}, and military target recognition \cite{ref4}.

\begin{figure}[!t]
\centering
\includegraphics[width=0.5\textwidth]{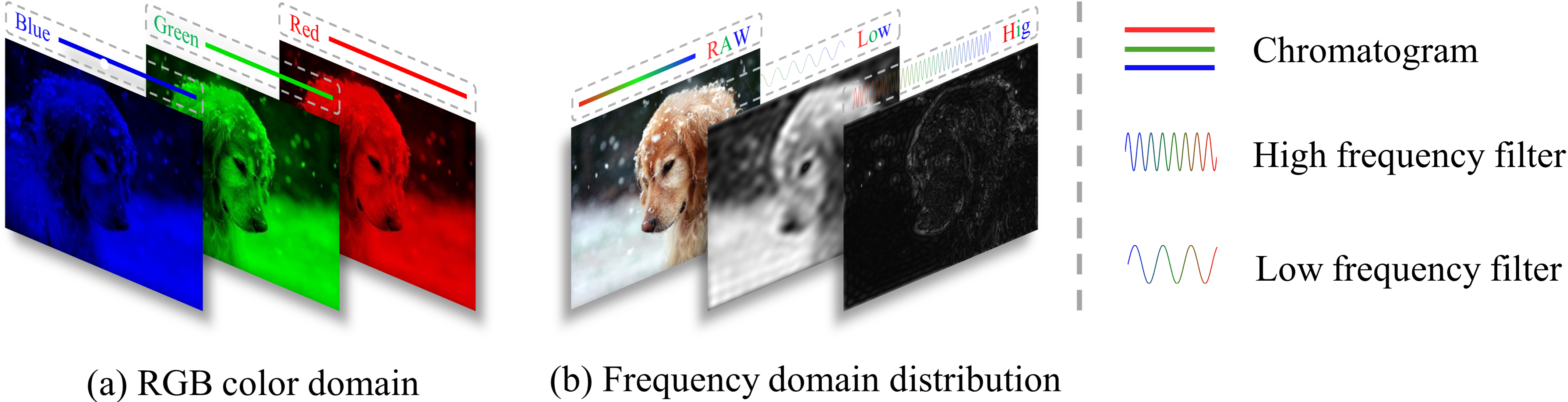}
\caption{Image RGB domain and frequency domain distribution display. (a) RGB color model. The three primary colors can provide information such as the color and texture of an image. (b) Frequency domain distribution. High-frequency distribution information represents rapid changes in the image and can provide image detail information, while low-frequency distribution information represents slow changes in the image and can offer global structural information.}
\label{Fig.1}
\end{figure}

WSSS aims to provide pixel-level predictions for images and can be considered as a visually intensive task. There are four common weakly-supervised annotation methods: (1) Image-level labeling, providing only the category information of the image; (2) Bounding box labeling, offering rough positional information of objects; (3) Scribble labeling, marking the approximate object location with simple lines; (4) Point labeling, indicating key object positions with one or more points. Although image-level labeling does not provide object positional information, it offers the highest accessibility and lowest annotation cost, thus becoming the most prevalent standard setting in weakly-supervised semantic segmentation tasks. By contrast, FSS aims to predict various objects using a small number of samples, which is more in line with the characteristics of difficult-to-obtain and long-tail distributed data in realistic open scenarios. FSS typically consists of two stages: (1) The meta-training stage, where the model is trained on a base dataset with disjoint classes from the test stage; (2) The meta-testing stage, where the model can quickly adapt and generalize to few-shot tasks for unseen categories, thus segmenting novel objects from the new class. However, the annotations for the base dataset under FSS setting are still labor-intensive and costly.

In this work, we replace the traditional pixel-level labels in FSS task with more challenging image-level labels, namely weakly-supervised few-shot semantic segmentation (WFSS). Similar to the majority of existing WSSS or FSS network models, current research on WFSS primarily focuses on obtaining more precise seed class activation maps (CAM), also known as pseudo masks, or refining the object boundaries for clearer outlines \cite{citation23},\cite{citation24},\cite{citation25},\cite{citation26},\cite{citation27}. However, the intrinsic nature of WFSS tasks indicates that base-class data can only provide the network with limited (few-shot) and meager (image-level label) semantic information. Therefore, the key to solving WFSS tasks lies in how to obtain more support information within the aforementioned limited constraints. Recently, Chen et al.\cite{method_14,method_13} demonstrate that features in networks can further decouple frequency domain distribution information. Motivated by this, we resort to the frequency domain during meta-training to pursue a desirable semantic transformation and propose an adaptive frequency-aware network (AFANet) for WFSS.

As shown in \cref{Fig.1}(a), current network models often rely on the RGB color domain of images to localize and predict objects by providing information such as color, texture, and surface features. However, this representation method still fails to meet the requirements of WFSS tasks. Natural images can further be decomposed into spatial frequency components (\cref{Fig.1}(b)) \cite{citation1,citation2}. That is, low-frequency distribution information represents slow or smooth changes in the image, and high-frequency distribution information represents rapid changes or image details. The former can provide global features and overall structural information of the image, while the latter can provide local details and minor feature information. Chen et al. \cite{method_14} further demonstrate that the output feature maps of convolutional layers can also be viewed as a mixture of different frequency information, and these mixed feature maps can be decoupled through octave convolutions. Hence, we propose a cross-granularity frequency-aware module (CFM) to decouple the cross-granularity information features in the pyramid network, and further optimize the semantic spatial structural information through realignment. By realigning and optimizing semantic spatial structural information through CFM, AFANet effectively addresses the limitations of current network models in WFSS tasks, providing unparalleled detail and accuracy in object prediction and localization. For more details, see the ablation study  \textit{Robustness Analysis of AFANet}.

In addition, recognizing the limitations of current network models in meeting the requirements of WFSS tasks, researchers have explored alternative approaches. One promising option is leveraging cross-modal models such as CLIP (Contrastive Language-Image Pretraining) \cite{method_8}. Unlike traditional models that rely on RGB color domain information, CLIP learns from a vast number of image-text pairs available on the web through joint learning. This model not only captures intricate visual details but also understands textual context, making it particularly adept at understanding and interpreting complex visual scenes. However, current research on CLIP mostly focuses on optimizing prompts or generating seed CAM through offline learning \cite{citation6,citation7}. Despite CLIP's demonstrated strong zero-shot capability in identifying unseen classes through extensive experiments, the rough utilization of offline learning, relying solely on prior knowledge, still exhibits significant discrepancies with the data distribution of downstream tasks \cite{citation3,citation4},\cite{citation5}.

Motivated by this, we further propose a CLIP-guided spatial-adapter module (CSM). CSM can perform spatial domain adaptive transformations on CLIP's cross-modal textual information based on the semantic distribution characteristics of downstream tasks. Through such fine-tuning and adaptation, CSM can effectively address the disparity between prior knowledge and the data distribution of downstream tasks. To the best of our knowledge, this is the first time that CLIP engages in collaborative updates with downstream task network models in an online learning manner. This learning approach significantly enhances the model's adaptability to new tasks, achieving a tight integration between the model and the task, thereby achieving superior performance in practical applications.

\begin{figure*}[!t]
\centering
\includegraphics[width=\textwidth]{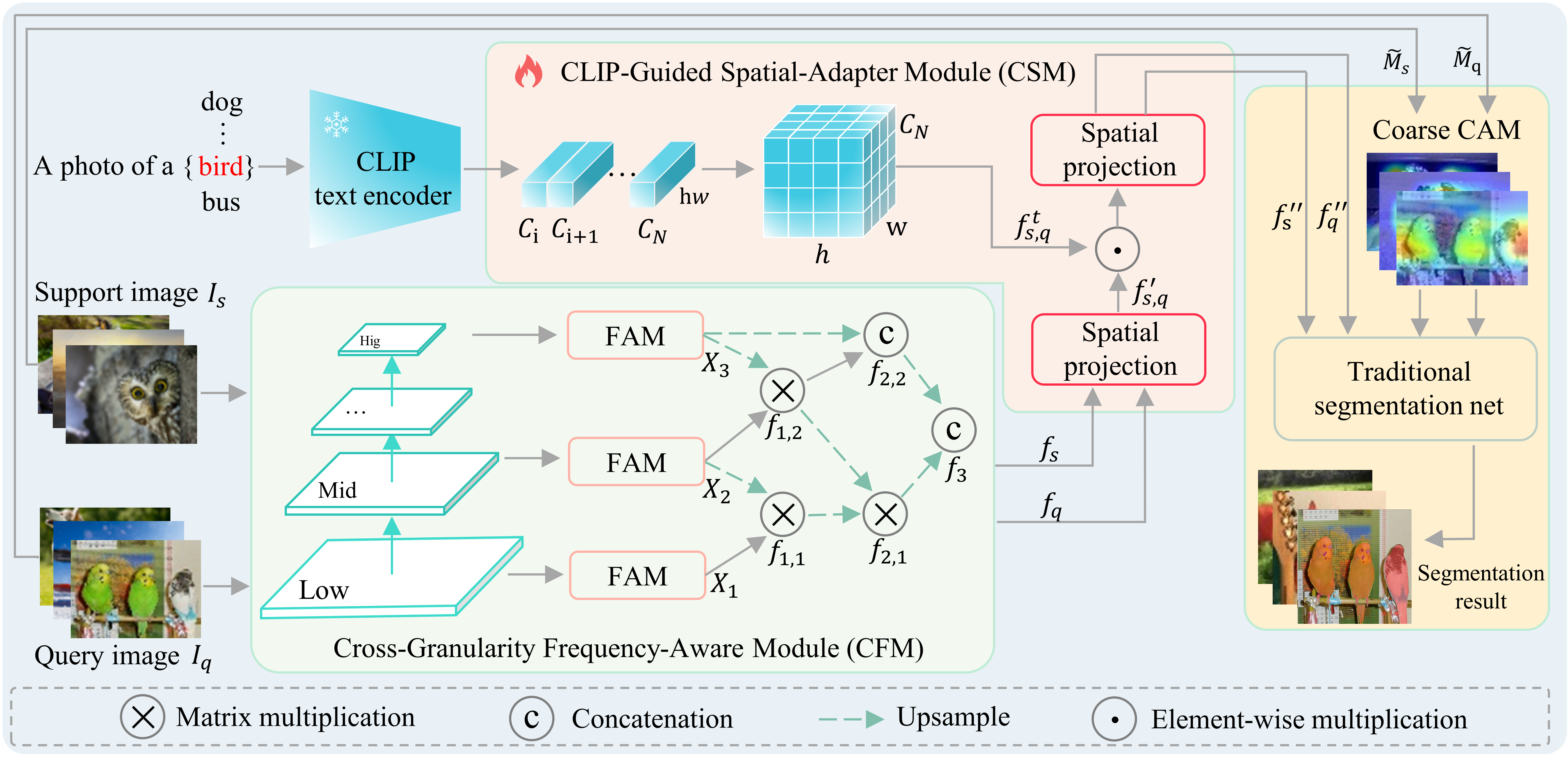}
\caption{\textbf{Overview of our AFANet framework.} (1) Cross-granularity frequency-aware module (CFM) extracts information from the low (layer 3), middle (layer 9) and high layers (layer 12) of the backbone respectively. Subsequently, FAM decomposes the RGB domain information into high-frequency and low-frequency distributions of different granularities and optimizes the spatial structural information of the frequency domain by realigning them. (2) CLIP-guided spatial-adapter module (CSM) first reshapes the CLIP text information to $C_{N}\times h \times w$ to adapt to the network model, and then, guided by the output $f_{s,q}$ of CFM, reduces the distribution gap between prior knowledge and the network model in a spatially adaptive manner. Finally, the CSM output information $f_{s}^{''} $ and $f_{q}^{''} $ are passed into the segmentation network together with the pseudo masks $\tilde{M}_{s}$ and $\tilde{M} _{q}$. Further details can be found in our baseline IMR-HSNet \cite{method_6}.}
\label{Fig.2}
\end{figure*}

To sum up, our contributions are: (1) We propose an Adaptive frequency-aware network (AFANet), which for the first time incorporates the frequency-domain distribution information of images into the WFSS task. Through the cross-granularity frequency-aware module, we demonstrate the effectiveness of providing frequency information other than RGB for WFSS. (2) We propose a CLIP-guided spatial-adapter module, which enables spatial domain adaptive transformations on CLIP's cross-modal textual information to adapt to the data distribution of downstream tasks. To the best of our knowledge, this is the first time that CLIP engages in collaborative updates with downstream task network models in an online learning manner. (3) Through extensive experiments and fair comparisons, AFANet has achieved state-of-the-art performance on the Pascal-5\textsuperscript{i} and COCO-20\textsuperscript{i} datasets.

The remainder of this paper is organized as follows. In Section~\ref{sec2}, we show some related works. In Section~\ref{sec3}, we illustrate the proposed framework elaborately. Section~\ref{sec4} reports the experimental results and presents the ablation studies. In Section~\ref{sec5}, we conclude this paper.

\section{Related Works}
\label{sec2}
\subsection{Few-shot Learning}

Few-shot learning aims to recognize novel classes using a limited number of examples, which are disjoint from the known classes \cite{citation32,citation33}. It is a fundamental paradigm for evaluating the data efficiency and adaptability of learning algorithms and has broad applications in the field of computer vision. Most existing few-shot learning methods can be categorized into transfer learning or meta-learning. Transfer learning typically involves pre-training models on diverse datasets and then fine-tuning them on tasks with limited data \cite{cit34, cit35}. The core idea of meta-learning is to acquire task-agnostic meta-knowledge from a large number of similar tasks and then utilize it to provide auxiliary guidance for few-shot tasks \cite{cit36,cit37}, \cite{cit38}. Meta-learning can be further divided into gradient-based methods and metric-based methods. Metric-based methods treat the metric space or metric strategy as meta-knowledge, typically used for learning image embeddings and comparing distances between them to derive highly correlated prototypes \cite{cit39}. Gradient-based methods consider the gradient optimization algorithm as meta-knowledge, enabling the model to quickly adapt to the characteristics of few-shot data distributions by updating hyperparameters \cite{cit40}, loss functions \cite{cit41}, and other parameters.

\subsection{Few-shot Semantic Segmentation}

Due to the necessity of extensive training data in traditional semantic segmentation methods \cite{fss1, fss2}, \cite{fss3, fss4}, the task of few-shot semantic segmentation has been proposed. This task aims to generate dense predictions for query images using a few annotated samples through meta-learning. Few-shot semantic segmentation methods are based on metric-based meta-learning and can be further divided into prototype-based guidance methods  \cite{fss5,fss6} or prior-knowledge-based guidance methods \cite{fss7, fss8}, \cite{fss9}. Prototype-based methods typically extract global \cite{fss10} or local features \cite{fss11} from images and then use multi-scale feature fusion  \cite{fss12} or cosine similarity \cite{fss13} to obtain semantic prototypes, which further guide the model in segmenting query images. Although prototype-based methods provide significant support for the model's general semantics, the prototype information obtained is ultimately limited due to the constraints of few-shot data. Consequently, relying solely on prototype-based methods may lead to reduced robustness of the segmentation results. To address this issue, methods based on prior knowledge have also been widely explored. This method can be further grouped into prior methods based on traditional knowledge or those based on large-language models. Prior methods based on traditional knowledge typically integrate inherent classifier knowledge \cite{fss14} or further extract semantic embeddings \cite{fss15} from the support and query sets, guiding the segmentation by either leveraging or suppressing the base class model knowledge \cite{fss16,fss17}. Prior methods based on large language models benefit from their powerful zero-shot semantic generalization capabilities, such as using CLIP or SAM to provide the model with extensive semantic prior information for guidance \cite{fss9,fss18}, \cite{fss19}. This method has significantly advanced the research progress in few-shot semantic segmentation and has also become a major area of focus in recent years.

\subsection{Weakly-Supervised Few-shot Segmentation}

Weakly-supervised few-shot semantic segmentation (WFSS) is closer to the practical scenarios in the real world, but the dual constraints of few-shot and weak supervision pose more severe challenges. There is only a small amount of relevant research. PANet and CANet \cite{method_1,method_2} have explored combining few-shot learning with weakly-supervised annotation methods. They replaced pixel-level labels with bounding boxes, but the model performance fell short of meeting the task requirements. Raza et al. \cite{method_3} proposed the first WFSS model using image-level labels, but this model only uses such labels during the testing phase, thus the WFSS task setting is not thorough. Siam et al. \cite{method_4} introduce multi-modal interaction information with word embeddings, reducing distribution discrepancy between visual and textual modalities by jointly guiding network attention with the visual module. Lee et al. \cite{method_5} generate CAM for each training class and determine their similarity by measuring the distance between the associated label's word embeddings. However, such word embedding space may lead to ineffective discrimination due to its lack of accuracy. Recently, IMR-HSNet \cite{method_6} has introduced the visual-language model CLIP to generate CAM as pseudo-masks for both support and query, refining masks through iterative interactions between each other. However, IMR-HSNet also relies on RGB domain information provided by images, thus having limited room for improvement under the constraints of few-shot and weak supervision. In contrast, our AFANet can overcome these limitations by offering new semantic support through spatial optimization in the frequency domain distribution.

\subsection{Contrastive Language-Image Pre-training (CLIP)}
Inspired by human cross-modal cognition \cite{method_7}, cross-modal learning aims to leverage data from other modalities to improve single-modal tasks. Recently, the vision-language pretraining model CLIP \cite{method_8} has been extensively trained on vision-language annotated pairs. By learning broad visual semantic concepts, CLIP has shown immense potential in zero-shot tasks and has become a milestone in the field of vision-language pretraining models \cite{citation8,citation9}. DenseCLIP \cite{method_9} first introduced CLIP into visually dense tasks, performing zero-shot object localization and segmentation in a context-aware manner. Besides, numerous works on optimizing CLIP prompts have been widely applied across various research fields. For example, researchers utilize the large-scale natural language model GPT-4 to optimize the category granularity of CLIP prompts \cite{method_10}, while CLIP-ES \cite{method_11} focuses on enhancing prompts for image foreground and background. However, these studies primarily provide prior information support through offline learning and may not fully adapt to downstream sub-tasks. To address these issues, in this work, we propose the CSM module, which collaboratively updates the downstream task network in an online learning manner, enhancing CLIP's adaptability to new tasks and achieving superior performance in practical applications.

\section{Methodology}
\label{sec3}
\subsection{Preliminary}
For a conventional 1-way K-shot few-shot semantic segmentation task, the common approach is to adopt a meta-learning paradigm. This involves setting up a base dataset \(D_{\text{base}}\) and a novel dataset \(D_{\text{novel}}\), where \(D_{base}\cap D_{novel}=\phi\). \(D_{\text{base}}\) contains a set of seen classes \(C_{\text{train}}\) used for training, while \(D_{\text{novel}}\) contains unseen classes used to evaluate the model's generalization ability to unseen categories. During training, \(D_\text{base}\) is often further partitioned into support set $\mathcal{S}$ and query set $\mathcal{Q}$, and multiple episodes are set up for periodic training using the metric learning method. It can be defined as

\begin{align}
  \mathcal{S}=\left\{\left(I_{s}^{k}, M_{s}^{k}\right)\right\}_{k=1}^{K}, \quad \mathcal{Q}=\left\{(I_{q}, M_{q}\right)\},
\end{align}
where, ${I_{s}^{k}\in \mathbb{R}^{C \times H \times W}}$ and ${M_{s}^{k}\in \mathbb{R}^{H \times W}}$ denote the input images and their corresponding binary masks, respectively. 

In the more challenging WFSS task, binary masks are replaced by image-level labels: 
\begin{align}
  \mathcal{S}=\left\{\left(I_{s}^{k}, C_{s}^{k}\right)\right\}_{k=1}^{K}, \quad \mathcal{Q}=\left\{(I_{q}, C_{q}\right)\}, 
\end{align}
where, \(C_{s}^{k}\) and $C_{q}$ represent the class labels for the support and query respectively.

However, image labels cannot be directly recognized by the segmentation network for model training. Therefore, it is necessary to further generate the CAM as pseudo-mask $\tilde{M}_{s}^{k}$ and $\tilde{M}_{q}$ corresponding to support \(\mathcal{S}\) and query $\mathcal{Q}$. For $\tilde{M}_{s}^{k}$, the CAM is computed by weighted summing the feature maps \( W_{s}^{k}\) of an image \(I_{s}^{k}\) with embedding class weights \(t_{i}^{k}\) obtained from the CLIP text encoder for class $C_{s}^{k}$. The relu function and max normalization are then applied to remove negative activations and scale the CAM to the range [0, 1]. It can be calculated as

\begin{align}
  \tilde{M}_{s}^{k}=\frac{{\text{relu}}({{t}_{i}^{k}}^{T}{W}_{s}^{k})}{\text{max}(\text{relu}({{t}_{i}^{k}}^{T}{W}_{s}^{k}))}.
\end{align}

During meta-testing, the trained model  $f(* | \theta)$ predicts the query mask  $\hat{M}_{q}$ using the given support images $I_{s}^{k}$ and the CAM  $\tilde{M}_{s}^{k}$ under \(D_{\text{novel}}\). Thus, the prediction is calculated as
\begin{align}
\hat{M}_{q} = f \left ( \left\{\left(I_{s}^{k}, \tilde{M}_{s}^{k}\right)\right\}_{k=1}^{K}, I_{q} | \theta \right ),
\end{align}
where, $f(* | \theta)$ is the trained model.

\subsection{Overview}

AFANet (\cref{Fig.2}) aims to introduce frequency-domain distribution and cross-modal text information adaptation to provide more semantic support for weakly-supervised few-shot semantic segmentation models. To extract more visual semantic information within limited constraints, the cross-granularity frequency-aware module (CFM) decouples the input RGB images $I_{s}$  and $I_{q}$ into high-frequency and low-frequency distributions. It then optimizes semantic structural information through realignment. Subsequently, the frequency-domain information guides CLIP, adapting the network model to cross-modal text information through online learning. Finally, the output features from CLIP-guided spatial-adapter module (CSM) are fed into a conventional segmentation network to achieve more accurate segmentation results. Detailed descriptions of each module of AFANet are provided in the following sections.

\subsection{Cross-Granularity Frequency-Aware Module (CFM)}
Compared to traditional RGB information, frequency domain information can provide more comprehensive information support. This viewpoint has also been validated in camouflage target detection tasks \cite{method_12,method_13}. As shown in \cref{Fig.1}, a common RGB image can also be decomposed into distributions of high-frequency and low-frequency. High-frequency information can reflect rapid changes in the image, such as foreground and background contours, thereby enhancing the global information of the image \cite{citation11,citation12}. On the other hand, low-frequency information changes more slowly and is coarser, typically appearing in the main parts of the image \cite{citation13,citation14}. In this work, we employ octave convolution \cite{method_14} to decompose the output features of convolutional layers into high-frequency and low-frequency distributions, providing neural networks with more comprehensive information support.

\begin{figure}[!t]
\centering
\includegraphics[width=0.5\textwidth]{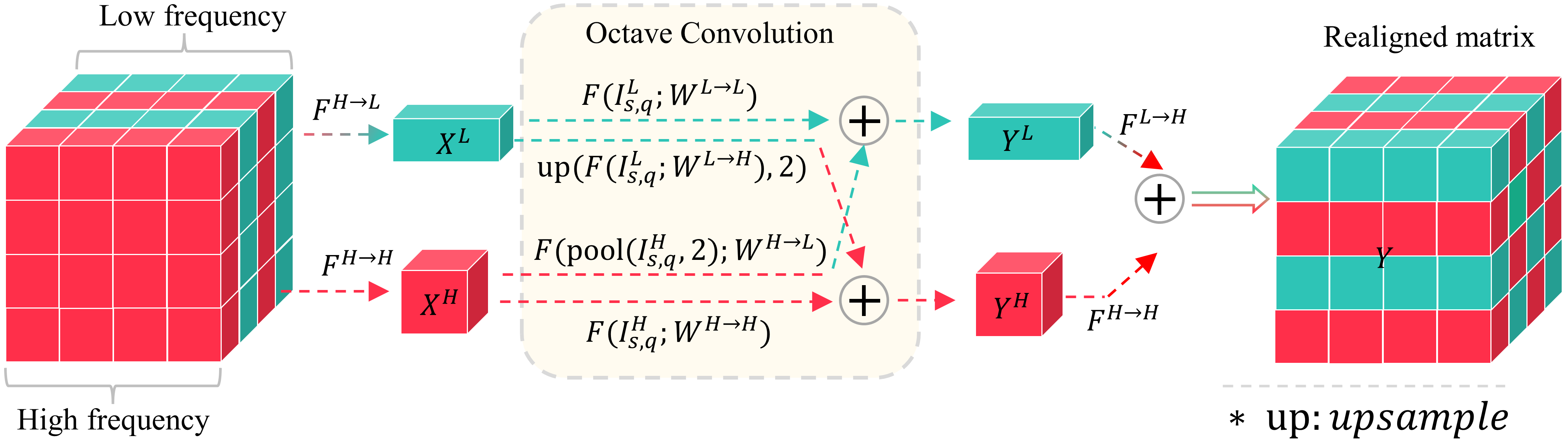}
\caption{Frequency-aware module (FAM). The green and red arrows denote the updating and exchange processes of low-frequency and high-frequency information, respectively.}
\label{Fig.3}
\end{figure}

Specifically, as shown in \cref{Fig.2}, given the input image support $I_{s}\in \mathbb{R}^{3 \times H \times W}$ and query $I_{q}\in \mathbb{R}^{3 \times H \times W}$, we first extract the features at different layers from the pyramid network \cite{method_15}, including low-level (layer 3), mid-level (layer 9), and high-level (layer 12) features. The combination of these features at different granularities enables a more comprehensive representation of both local details and global information in the image. Subsequently, as depicted in \cref{Fig.3}, the frequency-aware module (FAM) decouples these features at different granularities into high-frequency features $ \emph{F}^{H}$ and low-frequency features $ \emph{F}^{L}$ using octave convolution, and optimizes the frequency-domain structural information by realigning them. This process can be described as:

\begin{align}
Y_{s,q}^{H} &= F(I_{s,q}^{H};W^{H\to H}) + \text{Upsample}(F(I_{s,q}^{L};W^{L\to H}),2), \\
&\quad Y_{s,q}^{L} = F(I_{s,q}^{L};W^{L\to L}) + F(\text{pool}(I_{s,q}^{H},2);W^{H\to L}),
\end{align}

where, $F(I_{s,q};W)$ represents the octave convolution of input images $I_{s,q}$ and learnable parameters $W$, ${H\to H}$ indicate the preservation of high-frequency features, while ${L\to H}$ represents the process of converting low-frequency to high-frequency, and vice versa. Besides, $\text{pool}(I_{s,q}^{H},2)$ is an average pooling operation with a kernel size of $2 \times 2$, and $\text{upsample}(*, 2)$ is an upsampling operation using bilinear interpolation \cite{citation15} with a dilation rate of 2. At this point, the output feature $W_{s,q}$ from the convolutional layer has been decoupled into $\emph{F}^{H}$ and  $ \emph{F}^{L}$. Chen et al. \cite{method_14} argue that the low-frequency feature contains redundant information and thus discards it. Here, we empirically believe that the original convolutional feature $W_{s,q}$ exhibits partial overlap in spatial distribution between $\emph{F}^{H}$ and $ \emph{F}^{L}$, leading to information redundancy or misguidance. Therefore, simply realigning it can optimize the structural information in the frequency domain (see ablation study Table \ref{table3} for more details). As follows:

\begin{align}
F_{s,q}= \text{Resize}(Y_{s,q}^{H} \oplus  Y_{s,q}^{L}),
\end{align}
where, the resize adjusts $Y_{s,q}^{H}$ and $Y_{s,q}^{L}$ to a unified dimension, and $\oplus $ representing element-wise addition.

 As illustrated in \cref{Fig.3}, after passing through the FAM, we obtain frequency domain information from different network layers. However, due to the mixed characteristics of semantic information across cross-granularity and cross-frequency domains at this stage, simple linear addition cannot fully leverage the advantages of multi-angle semantic guidance. To address this issue, we utilize the neighbor connection decoder (NCD) \cite{method_15} to establish contextual correlations between frequency domain features across different layers. The process is as follows:
\begin{equation}
\left\{
\begin{aligned}
    f_{1,1} &= x_{1} \otimes g\uparrow (x_{2}), \\
    f_{1,2} &= x_{2} \otimes g\uparrow (x_{3}), \\
    f_{2,1} &= g\uparrow (f_{1,1}) \otimes g\uparrow (f_{1,2}), \\
    f_{2,2} &= cat{(f_{1,2} , g\uparrow (x_{3}))}, \\
    f_{3} &= cat{(g\uparrow(f_{2,1}), f_{2,2})}, 
\end{aligned}
\right.
\label{eq:your_label}
\end{equation}
where, $x_{1}$ is the low-level frequency domain feature, $x_{1} = F_{s,q}^{low}$, $x_{2}$ is the mid-level frequency domain feature, $x_{2} = F_{s,q}^{mid}$, $x_{3}$ is the hig-level frequency domain feature, $x_{3} = F_{s,q}^{hig}$. $g\uparrow (*)$ is an upsampling operation used to maintain feature dimension consistency. $\otimes$ represents element-wise multiplication.

\subsection{CLIP-Guided Spatial-Adapter Module (CSM)}
CLIP, a large-scale vision-language model, has been trained on a dataset of over 400 million image-text pairs, enabling it to learn a broad range of visual semantic concepts and demonstrating significant potential in zero-shot classification tasks. Currently, CLIP is widely applied to downstream tasks such as image classification \cite{citation16}, object detection \cite{citation17}, and semantic segmentation \cite{citation18},\cite{citation19},\cite{citation20},\cite{citation21},\cite{citation22}. However, these methods utilize CLIP's prior knowledge in an offline learning manner and may not fully adapt to downstream subtasks. In this work, we use frequency domain information as guidance and employ an online learning approach to adapt CLIP's textual prior knowledge to downstream tasks through spatial adaptation.

However, the text information \(t_{i}^{k}\in \mathbb{R}^{C_{N} \times hw}\) of CLIP's text encoder cannot directly adapt to the downstream network. Where, $C_{N} = N$ represents the number of categories, either Pascal-5\textsuperscript{i} (N=20) or COCO-20\textsuperscript{i} (N=80). Moreover, $hw = 1024$ represents the length of the feature vector $t_{i}^{k}$, we utilize a linear layer to reduce its dimensionality, 

\begin{align}
t_{i}^{k} = \text{Linear}(t_{i}^{k},hw),
\end{align}
where, $hw=625$. Then, we resize $t_{i}^{k}$ to the same size as the frequency domain information, denoted as 
\begin{align}
f_{s,q}^{t} = \text{Resize}(t_{i}^{k}),
\end{align}
where, $f_{s,q}^{t}$represents the CLIP textual feature vector aligned with the frequency domain feature dimensions, \(f_{s,q}^{t}\in \mathbb{R}^{C_{N} \times h \times w}\), $h=w=25$.

With that, we can proceed to match the frequency domain information $f_{s,q}$ and CLIP prior knowledge $f_{s,q}^{t}$. As mentioned above, CLIP is pretrained on large-scale datasets with comprehensive categories, while the frequency domain features are extracted from smaller-scale datasets like Pascal-5\textsuperscript{i} dataset or COCO-20\textsuperscript{i} dataset. This results in significant differences in feature distributions between them. Therefore, as shown in \cref{Fig.2}, we project $f_{s,q}$ into the spatial domain, utilizing bilinear interpolation to downsample its feature sizes from $50\times50$ to $25\times25$.

\begin{align}
f_{s,q}^{'}= \text{Proj}\downarrow (f_{s,q}),
\end{align}
where, $f_{*}^{'}$ represents the frequency domain features after downsampling, and $Proj\downarrow$ is the bilinear interpolation with a scaling factor of 0.5. This is because a smaller feature size implies a larger receptive field, achieved by removing redundant features, which further facilitates aligning the spatial feature distributions of $f_{s,q}^{'}$ and $f_{s,q}^{t}$.

Once the spatial feature distributions of them are aligned, we apply element-wise multiplication to enhance the feature representations. Then, we use bilinear interpolation again, but upsampling to increase the feature size from $25\times25$ to $50\times50$. 
\begin{align}
f_{s,q}^{''}= \text{Proj}\uparrow  (f_{s,q}^{'} \otimes f_{s,q}^{t}),
\end{align}
where $Proj\uparrow$ is the bilinear interpolation with a scaling factor of 2.

By upsampling, the network model AFANet can extract more fused feature details, enhancing the network's perceptual and expressive capabilities. This allows AFANet to capture semantic information in image segmentation tasks better, improving segmentation accuracy and performance.

\begin{figure*}[!t]
\centering
\includegraphics[width=0.9\textwidth]{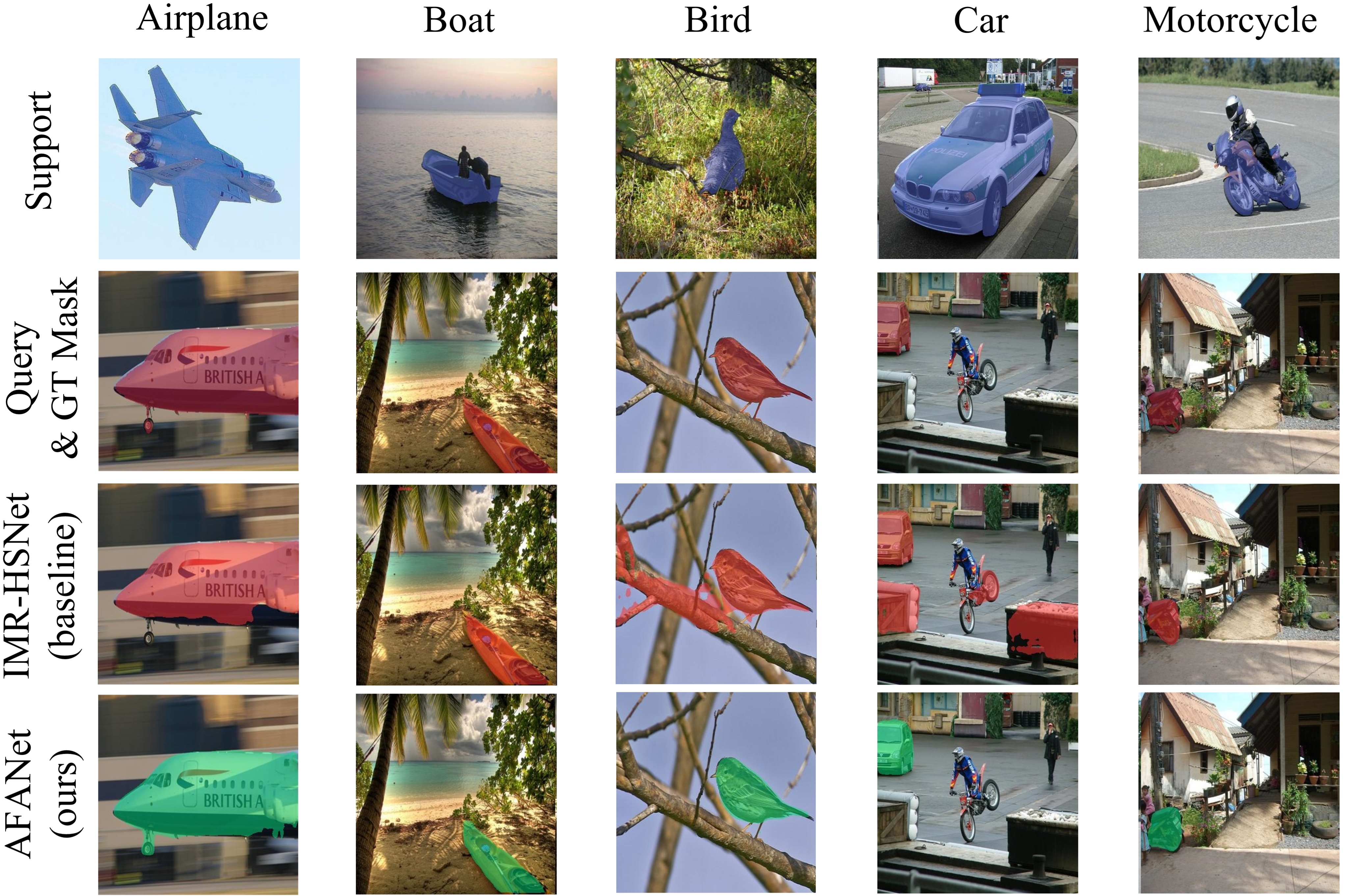}
\caption{Qualitative Analysis: Visualizing Segmentation Results Under a 1-Shot Setting. The data is sourced from Pascal-5\textsuperscript{i}. Organized top-down, each row represents the support image, query image (ground truth mask), segmentation results from IMR-HSNet (baseline), and segmentation results from our model (AFANet), respectively. Each column represents different categories.}
\label{Fig.4}
\end{figure*}

\section{Experiments}
\label{sec4}

\subsection{Training Loss for AFANet}
AFANet is trained in a meta-learning paradigm, and each episode is equivalent to a data sample in general learning algorithms. 
In segmentation, we using binary cross entropy (BCE) loss to update all parameters. In order to stay consistent with IMR-HSNet \cite{method_6}, we also add intermediate supervision to the output mask of each iteration. The final total loss is as follows:
\begin{align}
  \mathcal{L}_{all} =\sum_{t=1}^{N} \left [  \alpha\cdot  BCE(\hat{M}_{s}^{t},\tilde{M}_{s})+\beta \cdot BCE(\hat{M}_{q}^{t},\tilde{M}_{q}) \right ],
  \label{loss14}
\end{align}
where, $\hat{M}_{s}^{t}$ and $\tilde{M}_{s}$ represent the predicted mask and pseudo ground-truth mask, respectively. $\hat{M}_{q}^{t}$ and $\tilde{M}_{q}$ are not reiterated. Besides, N is the number of iterations, $\alpha$ and $\beta$ are hyperparameters.

\subsection{Datasets and Evaluation Metrics}

For a fair comparison, our experimental setup remains consistent with IMR-HSNet, differing only in the replacement of ground-truth masks with image-level labels compared to the conventional FSS setup. The datasets used to evaluate the model include Pascal-5\textsuperscript{i} \cite{data_1} and COCO-20\textsuperscript{i} \cite{data_2}. Pascal-5\textsuperscript{i} contains 20 categories, divided into 4 folds, each fold consists of 5 categories, 3 folds are used for training, and the remaining 1 fold is used for testing. COCO-20\textsuperscript{i} contains more than 80,000 images with 80 categories, making it a more challenging dataset. We also set it to 4 folds, each fold containing 20 categories. The training and testing methods are the same as Pascal-5\textsuperscript{i}. The mean intersection over union (mIoU) serves as our standard metric for evaluating model performance. Specifically, the IOU of each category is calculated based on the confusion matrix, and then the mIOU is obtained by averaging the IOU of each category.

\subsection{Implementation Details}

During both the training and testing phases, we set the input image size to $400\times400$. For Pascal-5\textsuperscript{i} and COCO-20\textsuperscript{i}, we respectively employ ResNet50 and VGG16 as backbones and set the number of epochs to 35. The difference lies in the training configurations: during Pascal-5\textsuperscript{i} training, the batch size is 16 with a learning rate of 4e-4, while during  COCO-20\textsuperscript{i} training, the batch size is 20 with a learning rate of 1e-4. For meta-testing under each cross-validation for the two datasets, we randomly sample 1,000 episodes (support-query pairs) from the test set and evaluate their metrics in 5-shot setting. All experiments are conducted with four RTX 3090 GPUs.

\subsection{Comparison with State-of-the-art}

In this section, we quantitatively and qualitatively compare AFANet with state-of-the-art weakly-supervised and conventional few-shot segmentation methods under the same evaluation metrics. These metrics are evaluated on the Pascal-5\textsuperscript{i} and COCO-20\textsuperscript{i} datasets, respectively.

\begin{table*}[]
\centering
\caption{Comparisons of regular and weakly-supervised few-shot semantic segmentation methods on Pascal-5\textsuperscript{i}.}
\scalebox{0.95}{
\begin{tabular}{cc|l|c|p{1cm}cccc|p{1cm}cccc}

\toprule
                            & \multirow{3}{*}{Backbone}  \quad & \multicolumn{1}{c|}{\multirow{3}{*}{Methods}} & \multirow{3}{*}{A. Type} & \multicolumn{5}{c}{1-shot}        & \multicolumn{5}{c}{5-shot}         \\
                             \cmidrule(){5-14}         
                            &                            \quad &                          &                          \quad & \centering $5^{0}$    \quad & $5^{1}$   \quad  & $5^{2}$    \quad & $5^{3}$    \quad & Mean  \quad & \centering $5^{0}$    \quad & $5^{1}$     \quad & $5^{2}$     \quad & $5^{3}$    \quad & Mean \\
                            \cmidrule(){1-14}                  
                            & \multirow{8}{*}{VGG16}     \quad & RPMG (TCSVT 23) \cite{rpmg}          & \textit{P}               \quad & \centering 58.3 \quad & 69.9 \quad & 55.7 \quad & 53.8 \quad & 59.4  \quad & \centering 59.9 \quad & 70.8  \quad & 55.4  \quad & 56.8 \quad & 60.7 \\
                            &                            \quad & HDMNet (CVPR 23) \cite{hdmnet}        & \textit{P}              \quad & \centering 64.8 \quad & 71.4 \quad & 67.7 \quad & 56.4 \quad & 65.1  \quad & \centering 68.1 \quad & 73.1  \quad & 71.8  \quad & 64.0 \quad & 69.3 \\
                            &                            \quad  & PFENet++ (TPAMI 23) \cite{pfenet++}      & \textit{P}           \quad & \centering 59.2 \quad & 69.6 \quad & 66.8 \quad & 60.7 \quad & 64.1  \quad & \centering 64.3 \quad & 72.0  \quad & 70.0  \quad & 62.7 \quad & 67.3 \\
                            &                            \quad & DRNet (TCSVT 24) \cite{drnet}        & \textit{P}               \quad & \centering 65.4 \quad & 66.2 \quad & 54.2 \quad & 53.8 \quad & 59.9  \quad & \centering 68.9 \quad & 68.2  \quad & 64.6  \quad & 57.1 \quad & 64.7 \\
                            \cmidrule(){3-14}  
                            &                           \quad  & PANet (ICCV 19) \cite{method_1}          & \textit{B}           \quad & \centering -    \quad & -    \quad & -    \quad & -    \quad & 45.1  \quad &\centering  -    \quad & -     \quad & -     \quad & -    \quad & 52.8 \\
                            \cmidrule(){3-14} 
                            &                            \quad & Pix-MetaNet (WACV 22) \cite{method_5}   & \textit{I}            \quad & \centering 36.5 \quad & 51.7 \quad & 45.9 \quad & 35.6 \quad & 42.4  \quad & \centering -    \quad & -     \quad & -     \quad & -    \quad & -    \\
                            &                            \quad & IMR-HSNet (IJCAI 22) \cite{method_6}    & \textit{I}            \quad & \centering 58.2 \quad & 63.9 \quad & 52.9 \quad & 51.2 \quad & 56.5  \quad & \centering 60.5 \quad & 65.3  \quad & 55   \quad & 51.7  \quad & 58.1 \\
                            &                            \quad & \textbf{\text{AFANet (Ours)}}             & \textit{I}           \quad & \centering \textcolor{red}{64.1} \quad & \textcolor{red}{65.9} \quad & \textcolor{red}{59.2} \quad & \textcolor{red}{56.9} \quad & ${\color{red} 61.5^{\uparrow 5}}$  \quad & \centering \textcolor{red}{66.3} \quad & \textcolor{red}{68.7}  \quad & \textcolor{red}{61.6} \quad & \textcolor{red}{59.7} \quad & ${\color{red} 64.1^{\uparrow 6}}$ \\
                            \cmidrule(){2-14}  
                            & \multirow{12}{*}{Resnet50} \quad & DPNet (AAAI 22) \cite{dpnet}         & \textit{P}               \quad & \centering 60.7 \quad & 69.5 \quad & 62.8 \quad & 58.0 \quad & 62.7  \quad & \centering 64.7 \quad & 70.8  \quad & 69    \quad & 60.1 \quad & 66.2 \\
                            &                            \quad & RPMG (TCSVT 23)   \cite{rpmg}       & \textit{P}                \quad & \centering 64.4 \quad & 72.6 \quad & 57.9 \quad & 58.4 \quad & 63.3  \quad & \centering 65.3 \quad & 72.8  \quad & 58.4  \quad & 59.8 \quad & 64.1 \\
                            &                            \quad & HDMNet (CVPR 23) \cite{hdmnet}       & \textit{P}               \quad & \centering 71.0 \quad & 75.4 \quad & 68.9 \quad & 62.1 \quad & 69.4  \quad & \centering 71.3 \quad & 76.2  \quad & 71.3  \quad & 68.5 \quad & 71.8 \\
                            &                            \quad & PFENet++ (TPAMI 23) \cite{pfenet++}     & \textit{P}            \quad & \centering 60.6 \quad & 70.3 \quad & 65.6 \quad & 60.3 \quad & 64.2  \quad & \centering 65.2 \quad & 73.6  \quad & 74.1  \quad & 65.3 \quad & 69.6 \\
                            &                            \quad & DRNet (TCSVT 24) \cite{drnet}        & \textit{P}               \quad & \centering 66.1 \quad & 68.8 \quad & 61.3 \quad & 58.2 \quad & 63.6  \quad & \centering 69.2 \quad & 73.9  \quad & 65.4  \quad & 65.3 \quad & 68.5 \\
                            \cmidrule(){3-14}
                            &                            \quad & CANet (CVPR 19)  \cite{method_2}        & \textit{B}            \quad & \centering -    \quad & -    \quad & -    \quad & -    \quad & 52.0  \quad & \centering -    \quad & -     \quad & -     \quad & -    \quad & -    \\
                            \cmidrule(){3-14}
                            &                            \quad & Siam et al. (IJCAI 20) \cite{method_4}  & \textit{I}             \quad & \centering 49.5 \quad & 65.5 \quad & 50.0 \quad & 49.2 \quad & 53.5  \quad & \centering -    \quad & -     \quad & -     \quad & -    \quad & -    \\
                            &                            \quad & Zhang et al. (TMM 22) \cite{tmm}   & \textit{I}                  \quad & \centering 56.9 \quad & 62.5 \quad & 60.3 \quad & 49.9 \quad & 57.4  \quad & \centering -    \quad & -     \quad & -     \quad & -    \quad & -    \\
                            &                            \quad & IMR-HSNet (IJCAI 22) \cite{method_6}    & \textit{I}             \quad & \centering 62.6 \quad & \textcolor{red}{69.1} \quad & 56.1 \quad & 56.7 \quad & 61.0  \quad & \centering 63.6 \quad & 69.6  \quad & 56.3  \quad & 57.4 \quad & 61.8 \\
                            &                            \quad & MIAPnet (ICME 23) \cite{miapnet}       & \textit{I}              \quad & \centering 41.7 \quad & 51.3 \quad & 42.2 \quad & 41.8 \quad & 44.2  \quad & \centering -    \quad & -     \quad & -     \quad & -    \quad & -    \\
                            &                            \quad & \textbf{\text{AFANet (Ours)}}   & \textit{I}         \quad & \centering \textcolor{red}{65.7} \quad & 68.5 \quad &  \textcolor{red}{60.6} \quad &  \textcolor{red}{61.5} \quad &  ${\color{red} 64.0^{\uparrow 3}}$ \quad &  \centering \textcolor{red}{69.0} \quad &  \textcolor{red}{70.4} \quad &  \textcolor{red}{61.3} \quad &  \textcolor{red}{64.0} \quad &  ${\color{red} 66.2^{\uparrow 4.4}}$  \\
                            \bottomrule
\end{tabular}                                                   
}
\label{table1}

\raggedright Here, \textit{P}, \textit{B}, and \textit{I} denote pixel-level labels, bounding box labels, and image-level labels, respectively.  ${\color{red} \uparrow }$ represents the value of improvement from the baseline IMR-HSNet. The best mIoU obtained from weakly-supervised methods is highlighted in \textcolor{red}{red}.
\end{table*}

$\mathbf{Pascal-5\textsuperscript{i}.}$ The results under 1-shot and 5-shot settings, using VGG16 and ResNet50 as backbones respectively, are shown in Table \ref{table1}. We conclude that: (1) Regardless of the backbone or shot settings, AFANet achieves overwhelming performance improvements compared to existing WFSS methods. (2) Compared to our baseline model IMR-HSNet (VGG16, 5-shot), the maximum mIOU improvement reaches up to $6\%$. Additionally, in the 1-shot setting, AFANet's performance (61.5) with VGG16 as the backbone even surpasses that of IMR-HSNet using ResNet50 (61.0). (3) Compared to other few-shot semantic segmentation methods, AFANet also outperforms certain pixel-level supervised models, such as RPMG and DPnet. The outstanding performance of AFANet in the field of FSS fully demonstrates its robustness and adaptability across various scenarios. This further substantiates that frequency-domain information can extract more semantic information than RGB, and significantly enhances the adaptability of CLIP to downstream tasks.

$\mathbf{COCO-20\textsuperscript{i}.}$ Similarly, we illustrate the results in Table \ref{table2}. It still can be seen clearly that AFANet, when using VGG16 and ResNet as backbones respectively, consistently achieves state-of-the-art results in the WFSS task, regardless of whether it is in a 1-shot or 5-shot setting. For example, when using VGG16 as the backbone in a 1-shot setting, AFANet achieves the highest increase in mean mIOU compared to IMR-HSNet, with an increase of up to $3.6\%$. Even in the 1-shot setting, AFANet outperforms most pixel-level supervised methods, such as RPMG, PFENet++, and DRNet, regardless of whether it uses VGG16 or ResNet50 as the backbone. This confirms that AFANet is able to fully utilize the frequency domain distribution of images to extract richer semantic information. Additionally, through online learning of CLIP's prior knowledge, it better adapts to downstream tasks. We hope that our model can provide some inspiration for future WFSS research.

$\mathbf{Visualization.}$ In this subsection, we conduct visual qualitative analysis based on diverse real-world task scenarios. (1) Dynamic object detection scenarios. As shown in the first column of Fig. \ref{Fig.4}, moving objects in imaging may appear blurred, and the degree of blur variation also contains frequency domain distribution information. Therefore, AFANet segments more complete contours of the airplane compared to IMR-HSnet. (2) Camouflage target detection. As shown in the second column, due to the similarity in color between the boat and the beach, it is easy to cause visual confusion. However, frequency domain information can somewhat mitigate visual deception \cite{method_12}. AFANet's more accurate segmentation of the upper left corner of the boat further confirms this point. (3) Associative background suppression. Lee et al. \cite{Lee} have shown that neural networks may misidentify the background of certain objects as foreground when processing certain images, a phenomenon known as correlated background. For example, birds often appear in trees (background), and fish often appear in water (background). As shown in the third column, IMR-HSNet incorrectly identifies the tree branch (background) as foreground, but AFANet correctly identifies it. We empirically believe that after fusion learning of frequency domain information and CLIP, AFANet has stronger global and local recognition capabilities. (4) Multiple object and occlusion detection. In urban road scenes, detecting multiple objects or occluded objects is one of the fundamental tasks in computer vision. As shown in the third and fourth columns, AFANet accurately identifies target objects in multi-target images, and it also provides more complete segmentation of occluded objects. Overall, through qualitative visual analysis, we have demonstrated that frequency domain distributions can provide richer information support for neural networks. Additionally, after online learning, the adaptation effect of CLIP to downstream tasks has significantly improved.

\begin{table*}[t]
\centering
\caption{Comparisons of regular and weakly-supervised few-shot semantic segmentation methods on COCO-20\textsuperscript{i}.}
\scalebox{0.95}{
\begin{tabular}{cc|l|c|p{1cm}cccc|p{1cm}cccc}

\toprule
                            & \multirow{3}{*}{Backbone}  \quad & \multicolumn{1}{c|}{\multirow{3}{*}{Methods}} & \multirow{3}{*}{A. Type} & \multicolumn{5}{c}{1-shot}        & \multicolumn{5}{c}{5-shot}         \\
                                                                                                               \cmidrule(){5-14}                  
                            &                            \quad &                          &                          \quad & \centering $5^{0}$  \quad & $5^{1}$   \quad  & $5^{2}$    \quad & $5^{3}$    \quad & \centering Mean  \quad & \centering $5^{0}$    \quad & $5^{1}$     \quad & $5^{2}$     \quad & $5^{3}$    \quad & Mean \\
                            \cmidrule(){1-14}                 
                            & \multirow{8}{*}{VGG16}     \quad & RPMG (TCSVT 23)          & \textit{P}               \quad & \centering 34.5     \quad & 36.9 \quad & 35.3 \quad & 33.7 \quad & \centering 35.1  \quad & \centering -    \quad & -     \quad & -     \quad & -    \quad & -  \\
                            &                            \quad & HDMNet (CVPR 23)         & \textit{P}               \quad & \centering 40.7     \quad & 50.6 \quad & 48.2 \quad & 44.0 \quad & \centering 45.9  \quad & \centering 47.0 \quad & 56.5  \quad & 54.1  \quad & 51.9 \quad & 52.4 \\
                            &                           \quad  & PFENet++ (TPAMI 23)      & \textit{P}               \quad & \centering 38.6     \quad & 43.1 \quad & 40.0 \quad & 39.5 \quad & \centering 40.3  \quad & \centering 38.9 \quad & 46.0  \quad & 44.2  \quad & 44.1 \quad & 43.3 \\
                            &                            \quad & DRNet (TCSVT 24)         & \textit{P}               \quad & \centering 33.5     \quad & 33.4 \quad & 32.1 \quad & 34.2 \quad & \centering 33.3  \quad & \centering 43.1 \quad & 46.8  \quad & 40.9  \quad & 40.9 \quad & 42.9 \\
                            \cmidrule(){3-14}  
                            &                           \quad  & PANet (ICCV 19)          & \textit{B}               \quad & \centering 12.7     \quad & 8.7    \quad & 5.9    \quad & 4.8    \quad & \centering 8.0  \quad & \centering -    \quad & -     \quad & -     \quad & -    \quad & - \\
                            \cmidrule(){3-14} 
                            &                            \quad & Pix-MetaNet (WACV 22)    & \textit{I}               \quad & \centering 24.2     \quad & 12.9 \quad & 17.0 \quad & 14.0 \quad & \centering 17.0  \quad & \centering -    \quad & -     \quad & -     \quad & -    \quad & 17.5    \\
                            &                            \quad & IMR-HSNet (IJCAI 22)     & \textit{I}               \quad & \centering 34.9     \quad & 38.8 \quad & 37.0 \quad & 40.1 \quad & \centering 37.7  \quad & \centering 34.8 \quad & 41.0  \quad & 37.2   \quad & 39.7  \quad & 38.2 \\
                            &                \quad & \textbf{\text{AFANet (Ours)}}        & \textit{I}               \quad & \centering \textcolor{red}{38.3} \quad & \textcolor{red}{42.5} \quad & \textcolor{red}{42.9} \quad & \textcolor{red}{41.5} \quad & \centering ${\color{red} 41.3^{\uparrow 3.6}}$  \quad & \centering \textcolor{red}{37.9} \quad & \textcolor{red}{42.7}  \quad & \textcolor{red}{40.6} \quad & \textcolor{red}{43.1} \quad & ${\color{red} 41.1^{\uparrow 2.9}}$ \\
                            \cmidrule(){2-14}                              
                            & \multirow{12}{*}{Resnet50} \quad & RPMG (TCSVT 23)          & \textit{P}               \quad & \centering 38.3     \quad & 41.4 \quad & 39.6 \quad & 35.9 \quad & \centering 38.8  \quad & \centering -    \quad & -     \quad & -     \quad & -    \quad & - \\
                            &                            \quad & HDMNet (CVPR 23)         & \textit{P}               \quad & \centering 43.8     \quad & 55.3 \quad & 51.6 \quad & 49.4 \quad & \centering 50.0  \quad & \centering 50.6 \quad & 61.6  \quad & 55.7  \quad & 56.0 \quad & 56.0 \\
                            &                            \quad & PFENet++ (TPAMI 23)      & \textit{P}               \quad & \centering 40.9     \quad & 44.8 \quad & 39.7 \quad & 38.8 \quad & \centering 41.0  \quad & \centering 45.7 \quad & 52.4  \quad & 49.1  \quad & 47.2 \quad & 48.6 \\
                            &                            \quad & DRNet (TCSVT 24)         & \textit{P}               \quad & \centering 42.1     \quad & 42.8 \quad & 42.7 \quad & 41.3 \quad & \centering 42.2  \quad & \centering 47.7 \quad & 51.7  \quad & 47.0  \quad & 49.3 \quad & 49.0 \\
                            \cmidrule(){3-14}
                            &                            \quad & CANet (CVPR 19)          & \textit{B}                \quad & \centering -       \quad & -    \quad & -    \quad & -    \quad & \centering 49.9  \quad & \centering -    \quad & -     \quad & -     \quad & -    \quad & 51.6    \\
                            \cmidrule(){3-14}
                            &                            \quad & Siam et al. (IJCAI 20)   & \textit{I}                \quad & \centering -       \quad & -    \quad & -    \quad & -    \quad & \centering 15.0  \quad & \centering -    \quad & -     \quad & -     \quad & -    \quad & 15.6    \\
                            &                            \quad & Zhang et al. (TMM 22)    & \textit{I}                \quad & \centering 33.3    \quad & 32.0   \quad & 29.2 \quad & 29.2 \quad & \centering 30.9  \quad & \centering -    \quad & -     \quad & -     \quad & -    \quad & -    \\
                            &                            \quad & IMR-HSNet (IJCAI 22)     & \textit{I}                \quad & \centering 39.5    \quad & 43.8 \quad & 42.4 \quad & 44.1 \quad & \centering 42.4  \quad & \centering 40.7 \quad & 46.0  \quad & \textcolor{red}{45.0}  \quad & 46.0 \quad & 44.4\\
                            &                            \quad & MIAPnet (ICME 23)        & \textit{I}                \quad & \centering 34.9    \quad & 23.4 \quad & 12.4 \quad & 18.3 \quad &\centering  22.2  \quad & \centering -    \quad & -     \quad & -     \quad & -    \quad & -    \\
                            &                            \quad & \textbf{\text{AFANet (Ours)}}   & \textit{I}         \quad & \centering \textcolor{red}{40.2} \quad & \textcolor{red}{45.1} \quad &  \textcolor{red}{44.0} \quad &  \textcolor{red}{45.1} \quad &  \centering ${\color{red} 43.6^{\uparrow 1.2}}$ \quad &  \centering \textcolor{red}{41.0} \quad &  \textcolor{red}{49.5} \quad &  43.0 \quad &  \textcolor{red}{46.9} \quad &  ${\color{red} 45.1^{\uparrow 0.7}}$ \\
                            \bottomrule
\end{tabular}                                                   
}
\label{table2}

\raggedright Here, \textit{P}, \textit{B}, and \textit{I} denote pixel-level labels, bounding box labels, and image-level labels, respectively.  ${\color{red} \uparrow }$ represents the value of improvement from the baseline IMR-HSNet. The best mIoU obtained from weakly-supervised methods is highlighted in \textcolor{red}{red}.
\end{table*}

\subsection{Ablation study}

Pascal-5\textsuperscript{i} is used to perform the following ablation studies to investigate the impact of different modules and adapter sizes in the AFANet framework.

$\textbf{Effects of different modules}.$ CFM and CSM are two crucial modules of AFANet, detailed in subsections 3.3 and 3.4, respectively. Table \ref{table3} shows the segmentation results under different folds. We can observe that, with the gradual addition of CFM and CSM modules, there has been a steady improvement in segmentation performance. Particularly, the addition of the CSM module led to a 3.3\% increase in mean mIOU (four folds). This further demonstrates the importance of online learning in facilitating the adaptation of CLIP to downstream tasks.

\begin{table}[H]
\caption{Ablation study on various modules of AFANet}
\centering
\begin{tabularx}{\columnwidth}{>{\centering\arraybackslash}p{1cm}*{2}{>{\centering\arraybackslash}p{1cm}}|*{4}{>{\centering\arraybackslash}X}c}
\toprule
Baseline & CFM & CSM & $5^{0}$ & $5^{1}$ & $5^{2}$ & $5^{3}$ & Mean \\
\midrule
\checkmark & & & 62.6 & 69.1 & 56.1 & 56.7 & 61 \\
\checkmark & \checkmark & & 64.7 & 67.2 & 57.7 & 57.9 & 61.9 \\
\checkmark & \checkmark & \checkmark & 65.7 & 68.5 & 60.6 & 62.2 & 64.3 \\
\bottomrule
\end{tabularx}
\label{table3}

\raggedright Optimal results obtained by combining CFM and CSM.

\end{table}

$\textbf{Effects of different adapter sizes}.$ The adapter size is an important parameter in the CSM for adjusting the size of $f_{s,q}$ and the fused feature resulting from the combination of $f_{s,q}^{'} and $ $f_{s,q}^{t}$. As described in subsection 3.3, a smaller adapter size can filter out redundant features of CLIP prior knowledge that are irrelevant to downstream tasks, while a larger adapter size can explore more semantic details after the fusion of CLIP and frequency domain features. As shown in the results of Table \ref{table4}, we compare different adapter sizes in a linearly increasing manner and ultimately select 25 as the final choice.

\begin{table}[H]
\caption{Ablation study on different adapter sizes.}
\centering
\begin{tabularx}{\columnwidth}{c|*{4}{>{\centering\arraybackslash}X}c}
\toprule
Adapter size & $5^{0}$ & $5^{1}$ & $5^{2}$ & $5^{3}$ & Mean \\
\midrule
20 & 64.4 & 67.7 & 60.0 & 60.5 & 63.2 \\
25 & 65.7 & 68.5 & 60.6 & 62.2 & 64.3 \\
50 & 63.8 & 68.0 & 60.3 & 61.8 & 63.5 \\
\bottomrule
\end{tabularx}
\label{table4}

\raggedright The optimal result is achieved when the adapter size is 25.

\end{table}

\textbf{Effects of different hyper-parameters $\alpha$ and $\beta$.} In all experiments, we set $\alpha$ and $\beta$ to 1 (Eq. (\ref{loss14})). We further vary their values between $\left \{0.4, 0.6, 1\right \}$  to select the most robust hyperparameters. As show in Table \ref{table5}, AFANet achieves the best results when both $\alpha$ and $\beta$ are set to 1.

\begin{table}[H]
\caption{Ablation study on different $\alpha$ and $\beta$ in loss function.}
\centering
\begin{tabularx}{\columnwidth}{>{\centering\arraybackslash}p{1cm} >{\centering\arraybackslash}p{1cm}| *{4}{>{\centering\arraybackslash}X} c}
\toprule
$\alpha $ & $\beta$ & 0 & 1 & 2 & 3 & Mean \\
\midrule
0.4 & 0.6 & 65.0 & 68.7 & 60.0 & 61.5 & 63.8 \\
0.6 & 0.4 & 65.0 & 68.1 & 60.1 & 62.0 & 63.8 \\
1.0 & 1.0 & 65.7 & 68.5 & 60.6 & 61.5 & 64.0 \\
\bottomrule
\end{tabularx}
\label{table5}
\end{table}

\textbf{Robustness Analysis of AFANet.} The backbone network can be divided into shallow and deep networks. Shallow networks are responsible for extracting low-level features such as object edges, while deep networks can build high-level semantic information.  Some previous studies have suggested that shallow features are more suitable for training models, whereas deep features are better for inference. Therefore, mid-level features are often used for feature extraction \cite{method_6, fss19}. However, recent studies have demonstrated that fixing the backbone weights and utilizing mid-level and high-level features from architectures like pyramid networks can significantly enhance the model's generalization ability \cite{fss15, fss20}.

\begin{figure}[H]  
\centering
\includegraphics[width=0.5\textwidth]{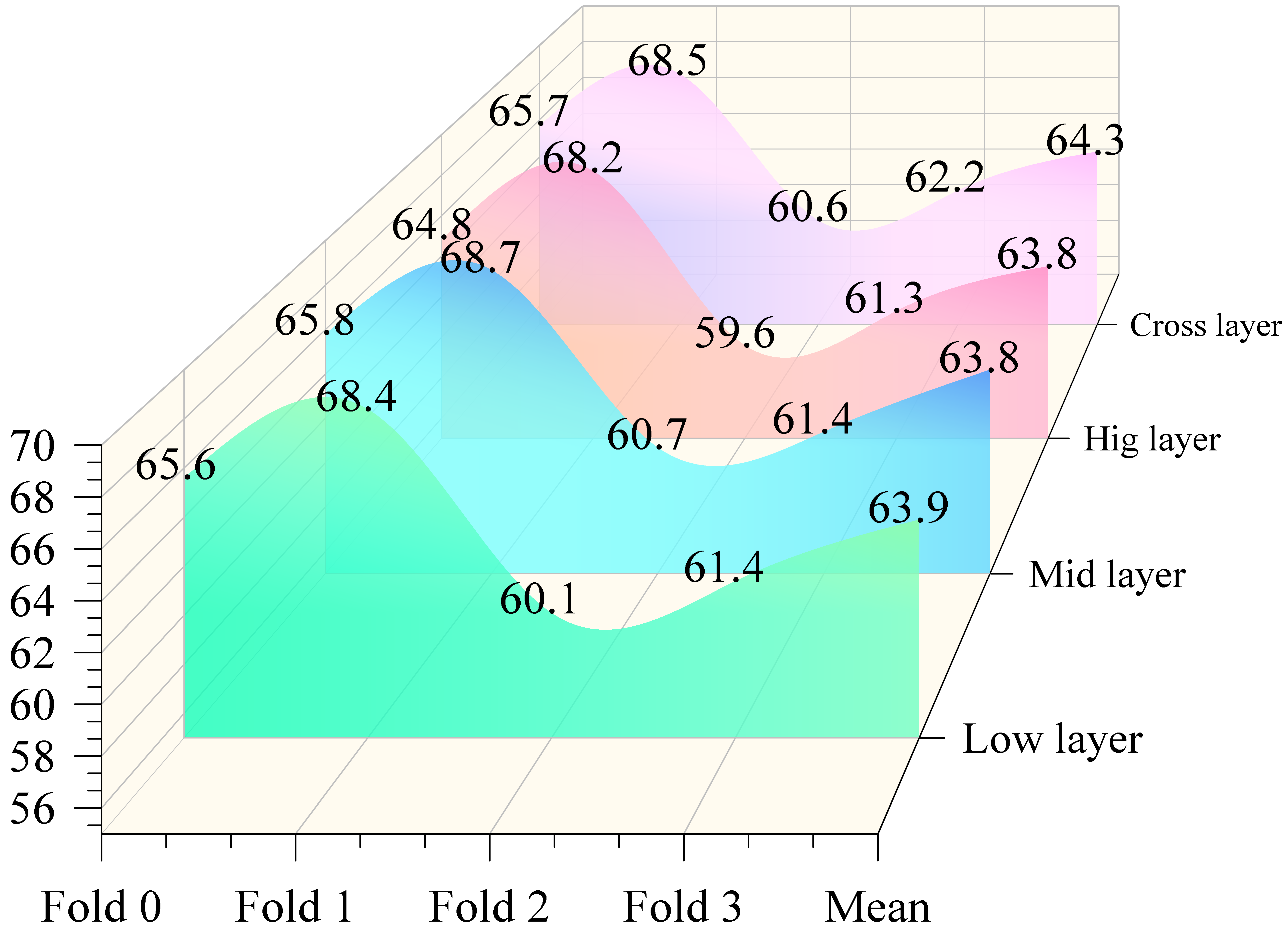}  
\caption{Ablation study of different network layers. Extract feature maps from a fixed pre-trained Backbone (ResNet-50) for low layers (0, 1, 2), mid layers (6, 7, 8), high layers (10, 11, 12), and our cross layers (3, 9, 12), respectively.}
\label{Fig.5}
\end{figure}

As shown in Fig. \ref{Fig.5}, we conduct ablation experiments on AFANet by extracting features from various layers of the backbone. Features extracted from the low layers (0, 1, 2), mid layers (6, 7, 8), and high layers (10, 11, 12) yield average mIoU scores of 63.9, 63.8, and 63.8, respectively. In contrast, our cross-layer features achieve the highest mIoU score of 64.3. These experimental results are consistent with previous studies, which indicate that using features from different network layers improves the model's generalization ability. However, it is noteworthy that the performance gain from cross-layer features compared to individual layers is not particularly significant. Furthermore, as illustrated in Fig. 5, AFANet performs highly consistently across different folds, regardless of which layer's features are utilized.

This robust observation aligns with the initial design intention of AFANet. The primary challenge in WFSS lies in the scarcity of data sources and information, making the introduction of additional semantic supervision crucial. The experiment further demonstrates that the frequency domain provides strong semantic guidance beyond RGB information, while CLIP, through online learning, further adapts to the data distribution characteristics of downstream tasks.

\section{Conclusion}
\label{sec5}
The adaptive frequency-aware network (AFANet) is proposed for tracking the WFSS task. AFANet consists of two modules: the cross-granularity frequency-aware module (CFM) and the CLIP-guided spatial-adapter module (CSM). CFM is used to extract the frequency distribution of RGB images, while CSM achieves adaptive integration of CLIP with downstream tasks through online learning. We hope this work can provide some inspiration to relevant researchers.

\bibliographystyle{IEEEtran}
\bibliography{my_tmm} %
\vfill

\end{document}